\documentclass[11pt]{article}
\usepackage{coling2018}
\usepackage{times}
\usepackage{url}
\usepackage{latexsym}

\usepackage{graphicx}
\usepackage{wrapfig}
\usepackage{url}
\usepackage{amsmath}
\usepackage{amssymb}
\usepackage{color,xcolor,colortbl}
\usepackage{enumitem}
\usepackage{algorithm}
\usepackage{algorithmic}
\usepackage[font={small}]{caption}
\usepackage{bm,bbm}
\usepackage{booktabs}
\usepackage{mathtools}
\usepackage{array}

\newcommand\numberthis{\addtocounter{equation}{1}\tag{\theequation}}

\definecolor{darkblue}{rgb}{0.0, 0.0, 0.55}
\newenvironment{fontppl}{\fontfamily{ppl}\selectfont}{\par} 

\title{Structure-Infused Copy Mechanisms for Abstractive Summarization}

\author{Kaiqiang Song \\
  Computer Science Dept.\\
  University of Central Florida \\
  Orlando, FL 32816, USA \\
  {\normalsize\tt kqsong@knights.ucf.edu} \\\And
  Lin Zhao \\
  Research and Tech. Center \\
  Robert Bosch LLC\\
  Sunnyvale, CA 94085, USA \\
  {\normalsize\tt lin.zhao@us.bosch.com} \\\And
  Fei Liu \\
  Computer Science Dept.\\
  University of Central Florida \\
  Orlando, FL 32816, USA \\
  {\normalsize\tt feiliu@cs.ucf.edu} \\}

\date{}

\begin{document}

\maketitle

\begin{abstract}

Seq2seq learning has produced promising results on summarization. 
However, in many cases, system summaries still struggle to keep the meaning of the original intact. 
They may miss out important words or relations that play critical roles in the syntactic structure of source sentences.
In this paper, we present structure-infused copy mechanisms to facilitate copying important words and relations from the source sentence to summary sentence. 
The approach naturally combines source dependency structure with the copy mechanism of an abstractive sentence summarizer.
Experimental results demonstrate the effectiveness of incorporating source-side syntactic information in the system, and our proposed approach compares favorably to state-of-the-art methods.

\end{abstract}

\section{Introduction}
\label{sec:introduction}

%
%
\blfootnote{
    %
    %
    %
    %
    %
    %
    \hspace{-0.65cm}  
    This work is licensed under a Creative Commons 
    Attribution 4.0 International License.
    License details:
    \url{http://creativecommons.org/licenses/by/4.0/}.
}

Recent years have witnessed increasing interest in abstractive summarization.
The systems seek to condense source texts to summaries that are concise, grammatical, and preserve the important meaning of the original texts~\cite{Nenkova:2011}.
The task encompasses a number of high-level text operations, e.g., paraphrasing, generalization, text reduction and reordering~\cite{Jing:1999},
posing a considerable challenge to natural language understanding.

\begin{table}[h]
\setlength{\tabcolsep}{5pt}
\renewcommand{\arraystretch}{1.1}
\centering
\begin{footnotesize}
\begin{fontppl}
\begin{tabular}{|l|l|}
\hline
\textbf{Src} & A Mozambican man suspect of murdering Jorge Microsse, director of Maputo central prison,\\
& has \textbf{\textcolor{red}{\emph{escaped}}} from the city's police headquarters, local media reported on Tuesday.\\
\textbf{Ref} & Mozambican suspected of killing Maputo prison director \textbf{\textcolor{red}{\emph{escapes}}}\\
\textbf{Sys} & mozambican man \textbf{\textcolor{red}{\emph{arrested}}} for murder\\
\hline
\hline
\textbf{Src} & An Alaska father who was too drunk to drive \textbf{\textcolor{red}{\emph{had}}} his 11-year-old son take the wheel, authorities said.\\
\textbf{Ref} & Drunk Alaska dad \textbf{\textcolor{red}{\emph{has}}} 11 year old drive home\\
\textbf{Sys} & alaska father who was too drunk to drive\\
\hline
\end{tabular}
\end{fontppl}
\end{footnotesize}
\caption{Example source sentences, reference and system summaries produced by a neural attentive seq-to-seq model.
System summaries fail to preserve summary-worthy content of the source (e.g., main verbs) despite their syntactic importance. 
}
\label{tab:example}
\vspace{-0.1in}
\end{table}

The sequence-to-sequence learning paradigm has achieved remarkable success on abstractive summarization~\cite{Rush:2015,Nallapati:2016,See:2017,Paulus:2017}.
While the results are impressive, individual system summaries can appear unreliable and fail to preserve the meaning of the source texts. 
Table~\ref{tab:example} presents two examples. 
In these cases, the syntactic structure of source sentences is relatively \emph{rare} but perfectly normal.
The first sentence contains two appositional phrases (``suspect of murdering Jorge Microsse,'' ``director of Maputo central prison'') and the second sentence has a relative clause (``who was too drunk to drive''), both located between the subject and the main verb.
The system, however, fails to identify the main verb in both cases; it instead chooses to focus on the first few words of the source sentences.
We observe that \emph{rare syntactic constructions} of the source can pose problems for neural summarization systems, possibly for two reasons.
First, similar to \emph{rare words}, certain syntactic constructions do not occur frequently enough in the training data to allow the system to learn the patterns. Second, neural summarization systems are not explicitly informed of the syntactic structure of the source sentences and they tend to bias towards sequential recency.

\begin{figure*}
\centering
\includegraphics[width=5.9in]{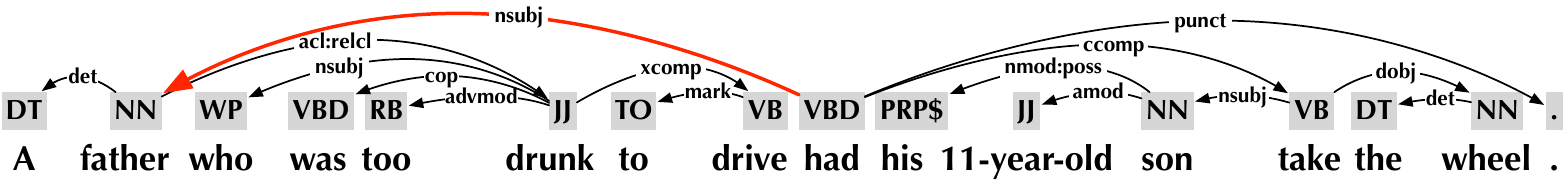}
\caption{An example dependency parse tree created for the source sentence in Table~\ref{tab:example}. 
If important dependency edges such as ``father $\leftarrow$ had'' can be preserved in the summary, the system summary is likely to preserve the meaning of the original.
}
\label{fig:example_dependency}
\vspace{-0.1in}
\end{figure*}

In this paper we seek to address this problem by incorporating source syntactic structure in neural sentence summarization to help the system identify summary-worthy content and compose summaries that preserve the important meaning of the source texts.
We present structure-infused copy mechanisms to facilitate copying source words and relations to the summary based on their semantic and structural importance in the source sentences. 
For example, if important parts of the source syntactic structure, such as a dependency edge from the main verb to the subject (``father'' $\leftarrow$ ``had,'' shown in Figure~\ref{fig:example_dependency}), 
can be preserved in the summary, the ``missing verb'' issue in Table~\ref{tab:example} can be effectively alleviated. 
Our model therefore learns to recognize important source words and source dependency relations and strives to preserve them in the summaries.
Our research contributions include the following:

\begin{itemize}[topsep=3pt,itemsep=-1pt,leftmargin=*]
\item we introduce novel neural architectures that encourage salient source words/relations to be preserved in summaries. The framework naturally combines the dependency parse tree structure with the copy mechanism of an abstractive summarization system. To the best of our knowledge, this is the first attempt at comparing various neural architectures for this purpose;

\item we study the effectiveness of several important components, including the vocabulary size, a coverage-based regularizer~\cite{See:2017}, and a beam search with reference mechanism~\cite{Tan:2017};

\item through extensive experiments we demonstrate that incorporating syntactic information in neural sentence summarization is effective. Our approach surpasses state-of-the-art published systems on the benchmark dataset.\footnote{We made our system publicly available at: \url{https://github.com/KaiQiangSong/struct_infused_summ}}
\end{itemize}

\section{Related Work}
\label{sec:related_work}

Prior to the deep learning era, sentence syntactic structure has been utilized to generate summaries with an ``\emph{extract-and-compress}'' framework.
Compressed summaries are generated using a joint model to extract sentences and drop non-important syntactic constituents~\cite{Daume:2002,Kirkpatrick:2011,Thadani:2013,Durrett:2016}, or a pipeline approach that combines generic sentence compression~\cite{McDonald:2006,Clarke:2008,Filippova:2015} with a sentence pre-selection or post-selection process~\cite{Zajic:2007,Galanis:2010,Wang:2013,Chen:2013,Chen:2014}.
Although syntactic information is helpful for summarization, there has been little prior work investigating how best to combine sentence syntactic structure with the neural abstractive summarization systems. 

Existing neural summarization systems handle syntactic structure only implicitly~\cite{Kikuchi:2016,Chen:2016,Zhou:2017,Tan:2017,Paulus:2017}.
Most systems adopt a ``\emph{cut-and-stitch}'' scheme that picks words either from the vocabulary or the source text and stitch them together using a recurrent language model. 
However, there lacks a mechanism to ensure structurally salient words and relations in source sentences are preserved in the summaries.
The resulting summary sentences can contain misleading information (e.g., ``mozambican man {arrested} for murder'' flips the meaning of the original) or grammatical errors (e.g., verbless, as in ``alaska father who was too drunk to drive'').

Natural language generation (NLG)-based abstractive summarization~\cite{Carenini:2008,Gerani:2014,Fabbrizio:2014,Liu:2015:NAACL,Takase:2016} also makes extensive use of structural information, including syntactic/semantic parse trees, discourse structures, and domain-specific templates built using a text planner or an OpenIE system~\cite{Pighin:2014}. 
In particular, Cao et al.~\shortcite{Cao:2018} leverage OpenIE and dependency parsing to extract fact tuples from the source
text and use those to improve the faithfulness of summaries.

Different from the above approaches, this paper seeks to directly incorporate source-side syntactic structure in the copy mechanism of an abstractive sentence summarization system.
It learns to recognize important source words and relations during training, while striving to preserve them in the summaries at test time to aid reproduction of factual details. 
Our intent of incorporating source syntax in summarization is different from that of neural machine translation (NMT)~\cite{Li:2017:NMT,Chen:2017:NMT}, in part because NMT does not handle the information loss from source to target. 
In contrast, a summarization system must selectively preserve source content to render concise and grammatical summaries.
We specifically focus on sentence summarization, where the goal is to reduce the first sentence of an article to a title-like summary.
We believe even for this reasonably simple task there remains issues unsolved.

\section{Our Approach}
\label{sec:framework}

We seek to transform a source sentence $\mathbf{x}$ to a summary sentence $\mathbf{y}$ that is concise, grammatical, and preserves the meaning of the source sentence.
A source word is replaced by its Glove embedding~\cite{Pennington:2014} before it is fed to the system; the vector is denoted by $\mathbf{x}_i$ ($i \in [S]$; `S' for source).
Similarly, a summary word is denoted by $\mathbf{y}_t$ ($t \in [T]$; `T' for target).
If a word does not appear in the input vocabulary, it is replaced by a special `{\fontfamily{fontppl}\selectfont $\langle\mbox{unk}\rangle$}' token.
We begin this section by describing the basic summarization framework, followed by our new copy mechanisms used to encourage source words and dependency relations to be preserved in the summary.

\subsection{The Basic Framework}

We build an encoder-decoder architecture for this work.
An encoder condenses the entire source text to a continuous vector;
it also learns a vector representation for each unit of the source text (e.g., words as units).
In this work we use a two-layer stacked bi-directional Long Short-Term Memory~\cite{Hochreiter:1997} networks as the encoder, where the input to the second layer is the concatenation of hidden states from the forward and backward passes of the first layer.
We obtain the hidden states of the second layer; they are denoted by $\mathbf{h}_i^e$.
The source text vector is constructed by averaging over all $\mathbf{h}_{i}^e$ and passing the vector through a feedforward layer with $\tanh$ activation to convert from the encoder hidden states to an initial decoder hidden state ($\mathbf{h}_0^d$).
This process is illustrated in Eq.~(\ref{equ:h_0_d}).
\begin{align*}
\displaystyle
& \mathbf{h}_i^e = f_e(\mathbf{h}_{i-1}^e, \mathbf{x}_i) \quad \mathbf{h}_t^d = f_d(\mathbf{h}_{t-1}^d, \mathbf{y}_{t-1})
\numberthis\label{equ:h_i_e}\\
& \mathbf{h}_0^d = \tanh(\mathbf{W}^{h_0} \frac{1}{S} \textstyle\sum\limits_{i=1}^S \mathbf{h}_i^e + \mathbf{b}^{h_0})
\numberthis\label{equ:h_0_d}
\end{align*}

A decoder unrolls the summary by predicting one word at a time.
During training, the decoder takes as input the embeddings of ground truth summary words, denoted by $\mathbf{y}_t$, while at test time $\mathbf{y}_t$ are embeddings of system predicted summary words (i.e., teacher forcing).
We implement an LSTM decoder with the attention mechanism.
A context vector $\mathbf{c}_t$ is used to encode the source words that the system attends to for generating the next summary word. 
It is defined in Eqs~(\ref{equ:e_t_i}-\ref{equ:c_t}),
where $[\cdot ||\cdot]$ denotes the concatenation of two vectors.
The $\boldsymbol\alpha$ matrix measures the strength of interaction between the decoder hidden states $\{\mathbf{h}_t^d\}$ and encoder hidden states $\{\mathbf{h}_i^e\}$.
To predict the next word, the context vector $\mathbf{c}_t$ and $\mathbf{h}_t^d$ are concatenated and used as input to build a new vector $\widetilde{\mathbf{h}}_t^d$ (Eq.~(\ref{equ:h_t_d_tilde})).
$\widetilde{\mathbf{h}}_t^d$ is a surrogate for semantic meanings carried at time step $t$ of the decoder.
It is subsequently used to compute a probability distribution over the output vocabulary (Eq.~(\ref{equ:p_vocab})).
\begin{align*}
\displaystyle
& {e}_{t,i} = \mathbf{v}^\top \tanh(\mathbf{W}^e [\mathbf{h}_t^d || \mathbf{h}_i^e] + b^e)
\numberthis\label{equ:e_t_i}\\
& \alpha_{t,i} = \frac{\exp(e_{t,i})}{\sum_{i'=1}^{S} \exp(e_{t,i'})}
\numberthis\label{equ:alpha_t_i}\\
& \mathbf{c}_t = \textstyle\sum_{i=1}^S \alpha_{t,i} \mathbf{h}_i^e
\numberthis\label{equ:c_t}\\
& \widetilde{\mathbf{h}}_t^d = \tanh(\mathbf{W}^h[\mathbf{h}_t^d || \mathbf{c}_t] + \mathbf{b}^h)
\numberthis\label{equ:h_t_d_tilde}\\
& P_{vocab}(w) = \text{softmax}(\mathbf{W}^y\widetilde{\mathbf{h}}_t^d + \mathbf{b}^y)
\numberthis\label{equ:p_vocab}
\end{align*}

The copy mechanism~\cite{Gulcehre:2016,See:2017} allows words in the source sequence to be selectively copied to the target sequence.
It expands the search space for summary words to include both the output vocabulary and the source text. 
The copy mechanism can effectively reduce out-of-vocabulary tokens in the generated text, potentially aiding a number of applications such as MT~\cite{Luong:2015:ACL} and text summarization~\cite{Gu:2016,Cheng:2016,Zeng:2017}.

Our copy mechanism employs a `switch' to estimate the likelihood of generating a word from the vocabulary ($p_{gen}$) vs. copying it from the source text ($1-p_{gen}$).  
The basic model is similar to that of the pointer-generator networks~\cite{See:2017}. 
The switch is a feedforward layer with sigmoid activation (Eq.~(\ref{equ:p_gen})). 
At time step $t$, its input is a concatenation of the decoder hidden state $\mathbf{h}_t^d$, context vector $\mathbf{c}_t$, and the embedding of the previously generated word $\mathbf{y}_{t-1}$.
For predicting the next word, we combine the generation and copy probabilities, shown in Eq.~(\ref{equ:p_w}). 
If a word $w$ appears once or more in the input text, its copy probability ($\sum_{i:w_i = w} \alpha_{t,i}$) is the sum of the attention weights over all its occurrences. 
If $w$ appears in both the vocabulary and source text, $P(w)$ is a weighted sum of the two probabilities.
{\medmuskip=1mu
\thinmuskip=1mu
\thickmuskip=1mu
\nulldelimiterspace=0pt
\scriptspace=0pt
\begin{align*}
p_{gen} &= \sigma(\mathbf{W}^z[\mathbf{h}_t^d || \mathbf{c}_t || \mathbf{y}_{t-1}]) + b^z)
\numberthis\label{equ:p_gen}\\
P(w) &= p_{gen} P_{vocab}(w) + (1 - p_{gen}) \sum_{i:w_i = w} \alpha_{t,i}
\numberthis\label{equ:p_w}
\end{align*}}

\vspace{-0.2in}
\subsection{Structure-Infused Copy Mechanisms}

The aforementioned copy mechanism attends to source words based on their ``semantic'' importance encoded in $\{\alpha_{t,i}\}$, which measures the semantic relatedness of the encoder hidden state $\mathbf{h}_i^e$ and the decoder hidden state $\mathbf{h}_t^d$ (Eq.~(\ref{equ:alpha_t_i})).
However, the source syntactic structure is ignored. 
This is problematic, because it hurts the system's ability to effectively identify summary-worthy source words that are syntactically important.
We next propose three strategies to inject source syntactic structure to the copy mechanism.

\begin{wraptable}{r}{0.5\textwidth}
\setlength{\tabcolsep}{4pt}
\renewcommand{\arraystretch}{1.1}
\centering
\begin{footnotesize}
\begin{fontppl}
\begin{tabular}{|l|c|}
\hline
\textbf{Structural info} & \textbf{Example}\\
\hline
(1) depth in the dependency parse tree & 0\\
(2) label of the incoming edge & `root'\\
(3) number of outgoing edges & 3\\
(4) part-of-speech tag & `VBD' \\
(5) absolution position in the source text & 9\\
(6) relative position in the source text & (0.5, 0.6] \\
\hline
\end{tabular}
\end{fontppl}
\end{footnotesize}
\caption{Six categories of structural labels.
Example labels are generated for word `had' in Figure~\ref{fig:example_dependency}.
Relative word positions are discretized into ten buckets.
}
\label{tab:structure}
\end{wraptable}

\begin{figure}
\centering
\includegraphics[width=4.2in]{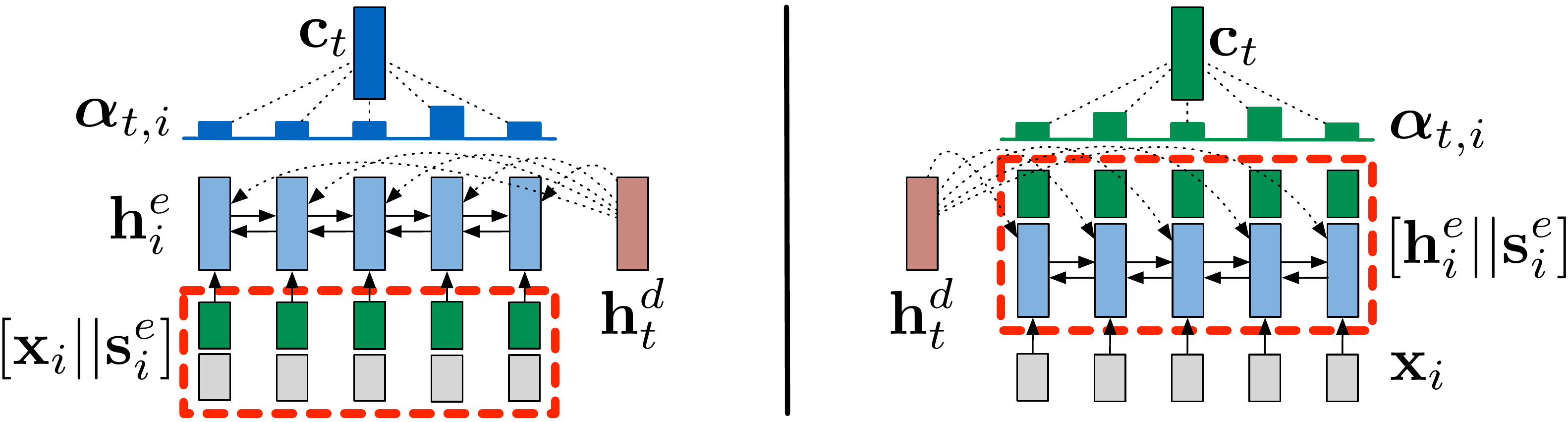}
\caption{System architectures for `Struct+Input' (left) and `Struct+Hidden' (right).
A critical question we seek to answer is whether the structural embeddings ($\mathbf{s}_i^e$) 
should be supplied as input to the encoder (left) or be exempted from encoding and directly concatenated with the encoder hidden states (right).
}
\label{fig:architecture_new}
\vspace{-0.1in}
\end{figure}

\subsubsection{Shallow Combination}

Inspired by compressive summarization via structured prediction~\cite{Kirkpatrick:2011,Almeida:2013}, we hypothesize that structural labels, such as the incoming dependency arc and the depth in a dependency parse tree, can be helpful to predict word importance.
We consider six categories of structural labels in this work; they are presented in Table~\ref{tab:structure}.
Each structural label is mapped to a fixed-length, trainable structural embedding.
However, a critical question remains as to where the structural embeddings should be injected in the existing neural architecture. 
This problem has not yet been systematically investigated.
In this work, we compare two settings: 
\begin{itemize}[topsep=3pt,itemsep=0pt]

\item \textbf{Struct+Input} concatenates structural embeddings of position $i$ (flattened into one vector $\mathbf{s}_i^e$) with the source word embedding $\mathbf{x}_i$ and uses them as a new form of input to the encoder:
$\mathbf{x}_i \Rightarrow [\mathbf{x}_i || \mathbf{s}_i^e]$;

\item \textbf{Struct+Hidden} concatenates structural embeddings of position $i$ (flattened) with the encoder hidden state $\mathbf{h}_i^e$ and uses them as a new form of hidden states: $\mathbf{h}_i^e \Rightarrow [\mathbf{h}_i^e || \mathbf{s}_i^e]$.

\end{itemize}

The architectural difference is illustrated in Figure~\ref{fig:architecture_new}.
Structural embeddings are important complements to existing neural architectures.
However, it is unclear whether they should be supplied as input to the encoder or be left out of the encoding process and directly concatenated with the encoder hidden states.
This is a critical question we seek to answer by comparing the two settings.
Note that an alternative setting is to separately encode words and structural labels using two RNN encoders, we consider this as a subproblem of the ``Struct+Input'' case.

The above models complement state-of-the-art by combining semantic and structural signals to determine summary-worthy content. 
Intuitively, a source word is copied to the summary for two reasons: it contains salient semantic content, or it serves a critical syntactic role in the source sentence.
Without explicitly modeling the two factors, `semantics' can outweigh `structure,' resulting in summaries that fail to keep the original meaning intact. 
In the following we propose a two-way mechanism to separately model the ``semantic'' and ``structural'' importance of source words.

\subsubsection{2-Way Combination (+Word)}

Our new architecture involves two attention matrices that are parallel to each other, denoted by $\boldsymbol\alpha$ and $\boldsymbol\beta$.
$\alpha_{t,i}$ is defined as previously in Eq.~(\ref{equ:e_t_i}-\ref{equ:alpha_t_i}).
It represents the ``semantic'' aspect, calculated as the strength of interaction between the encoder hidden state $\mathbf{h}_i^e$ and the decoder hidden state $\mathbf{h}_t^d$.
In contrast, $\beta_{t,i}$ measures the ``structural'' importance of the $i$-th input word to generating the $t$-th output word, calculated by comparing the structure-enhanced embedding $\mathbf{g}_i^e$ with the decoder hidden state $\mathbf{h}_t^d$ (Eq.~(\ref{equ:f_t_i}-\ref{equ:beta_t_i})).
We use $\mathbf{g}_i^e = [\mathbf{s}_i^e || \mathbf{x}_i]$ as a primitive (unencoded) representation of the $i$-th source word.

We define $\delta_{t,i} \propto \alpha_{t,i} + \epsilon\beta_{t,i}$ as a weighted sum of $\alpha_{t,i}$ and $\beta_{t,i}$, where a trainable coefficient $\epsilon$ is introduced to balance the contribution from both sides (Eq.~(\ref{equ:delta_t_i})).
Merging semantic and structural salience at this stage allows us to acquire an accurate estimate of how important the $i$-th source word is to predicting the $t$-th output word. 
$\delta_{t,i}$ replaces $\alpha_{t,i}$ to become the new attention value. 
It is used to calculate the context vector $\mathbf{c}_t$ (Eq.~(\ref{equ:c_t_new})).
A reliable estimate of $\mathbf{c}_t$ is crucial as it is used to estimate the generation probability over the vocabulary ($P_{vocab}(w)$, Eq.~(\ref{equ:h_t_d_tilde}-\ref{equ:p_vocab})), the switch value ($p_{gen}$, Eq.~(\ref{equ:p_gen})), and ultimately used to predict the next word ($P(w)$, Eq.~(\ref{equ:p_w})).
\begin{align*}
& f_{t,i} = \mathbf{u}^\top \tanh(\mathbf{W}^f[\mathbf{g}_i^e || \mathbf{h}_{t}^d] + \mathbf{b}^{f})
\numberthis\label{equ:f_t_i}\\
& \beta_{t,i} = \frac{\exp(f_{t,i})}{\sum_{i'=1}^S \exp(f_{t,i'})}
\numberthis\label{equ:beta_t_i}\\
& \delta_{t,i} = \frac{\alpha_{t,i} + \epsilon\beta_{t,i}}{\sum_{i'=1}^S(\alpha_{t,i'} + \epsilon\beta_{t,i'})}
\numberthis\label{equ:delta_t_i}\\
& \mathbf{c}_t = \sum_{i=1}^S \delta_{t,i} \mathbf{h}_i^e
\numberthis\label{equ:c_t_new}
\end{align*}

\begin{figure}[t]
\centering
\includegraphics[width=6in]{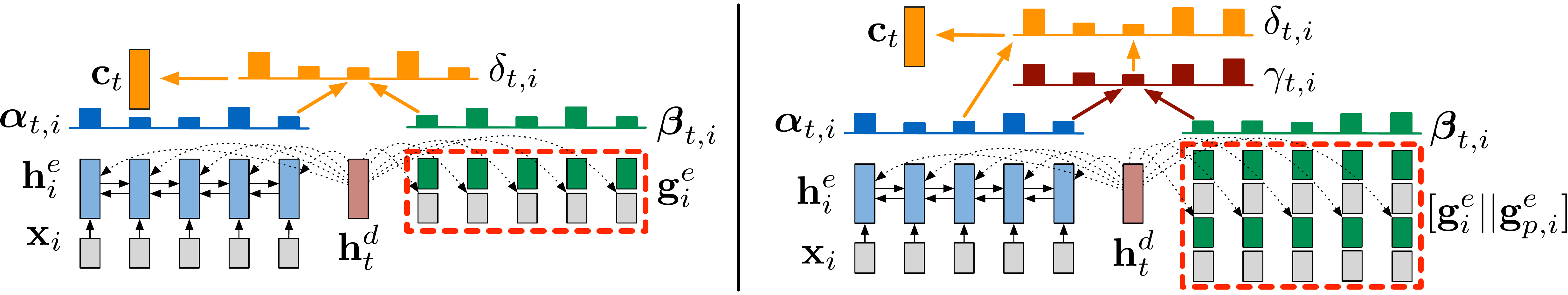}
\caption{System architectures for `Struct+2Way+Word' (left) and `Struct+2Way+Relation' (right).
$\beta_{t,i}$ (left) measures the structural importance of the $i$-th source word; 
$\beta_{t,i}$ (right) measures the saliency of the dependency edge pointing to the $i$-th source word.
$\mathbf{g}_{p,i}^e$ is the structural embedding of the parent.
In both cases $\delta_{t,i}$ replaces $\alpha_{t,i}$ to become the new attention value used to estimate the context vector $\mathbf{c}_t$.
}
\label{fig:architecture_2way}
\vspace{-0.1in}
\end{figure}

\subsubsection{2-Way Combination (+Relation)}

We observe that salient source relations also play a critical role in predicting the next word.
For example, if a dependency edge (``father'' $\xleftarrow{\vbox{\hbox{\scriptsize nsubj}\vskip-3pt}}$ ``had'') is salient and ``father'' is selected to be included in the summary, it is likely that ``had'' will be selected next such that a salient source relation (``nsubj'') is preserved in the summary.
Because summary words tend to follow the word order of the original, we assume selecting a source word and including it in the summary has an impact on its subsequent source words, but not the reverse.
\begin{figure}[h]
\centering
\includegraphics[width=2.8in]{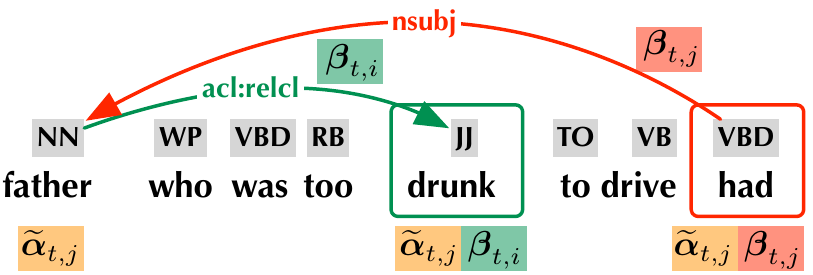}
\label{fig:relation}
\vspace{-0.1in}
\end{figure}

In this formulation we use $\beta_{t,i}$ to capture the saliency of the dependency edge pointing to the $i$-th source word.
Thus, an edge $w_j \leftarrow w_i$ has its salience score saved in $\beta_{t,j}$; 
and conversely, an edge $w_j \rightarrow w_i$ has its salience score in $\beta_{t,i}$.
$\boldsymbol\beta$ is calculated in the same way as described in Eq.~(\ref{equ:f_t_i}-\ref{equ:beta_t_i}).
However, we replace $\mathbf{g}_i^e$ with $[\mathbf{g}_i^e || \mathbf{g}_{p,i}^e]$ so that a dependency edge is characterized by the embeddings of its two endpoints ($\mathbf{g}_{p,i}^e$ is the parent embedding).
The architectural difference between ``Struct+2Way+Word'' and ``Struct+2Way+Relation'' is illustrated in Figure~\ref{fig:architecture_2way}.

To obtain the likelihood of $w_j$ being selected to the summary prior to time step $t$, we define $\widetilde{\alpha}_{t,j} = \sum_{t'=0}^{t-1} \alpha_{t',j}$ that sums up the individual probabilities up to time step $t$-1.
Assume there is a dependency edge $w_j \rightarrow w_i$ (j$<$i) whose salience score is denoted by $\beta_{t,i}$.
At time step $t$, we calculate  $\widetilde{\alpha}_{t,j}\beta_{t,i}$ (or $\widetilde{\alpha}_{t,j}\beta_{t,j}$ for edge $w_j \leftarrow w_i$)
as the probability of $w_i$ being selected to the summary, given that \emph{one} of its prior words $w_j$ (j$<$i) is included in the summary and there is a dependency edge connecting the two.
By summing the impact over \emph{all} its previous words, we obtain the likelihood of the $i$-th source word being included to the summary at time step $t$ in order to preserve salient source relations; this is denoted by $\gamma_{t,i}$ (Eq.~(\ref{equ:gamma_t_i})). 
Next, we define $\delta_{t,i} \propto \alpha_{t,i} + \epsilon\gamma_{t,i}$ as a weighted combination of semantic and structural salience (Eq.~(\ref{equ:delta_t_i_new})).
$\delta_{t,i}$ replace $\alpha_{t,i}$ to become the new attention values used to estimate the context vector $\mathbf{c}_t$ (Eq.~(\ref{equ:c_t_new})).
Finally, the calculation of generation probabilities $P_{vocab}(w)$, switch value $p_{gen}$, and probabilities for predicting the next word $P(w)$ remains the same as previously (Eq.~(\ref{equ:h_t_d_tilde}-\ref{equ:p_w})).
\begin{align*}
& \widetilde{\alpha}_{t,j} = \textstyle\sum_{t'=0}^{t-1} \alpha_{t',j}
\numberthis\label{equ:alpha_t_i_tilde}\\
& \gamma_{t,i} = \sum_{j:j<i}
\begin{cases}
\widetilde{\alpha}_{t,j}\beta_{t,i} \quad \mbox{if }w_j \rightarrow w_i\\
\widetilde{\alpha}_{t,j}\beta_{t,j} \quad \mbox{if }w_j \leftarrow w_i
\end{cases}
\numberthis\label{equ:gamma_t_i}\\
& \delta_{t,i} = \frac{\alpha_{t,i} + \epsilon\gamma_{t,i}}{\sum_{i'=1}^S(\alpha_{t,i'} + \epsilon\gamma_{t,i'})}
\numberthis\label{equ:delta_t_i_new}
\end{align*}

\vspace{0.1in}
\subsection{Learning Objective and Beam Search}
\label{sec:cov_beam}

We next describe our learning objective, including a coverage-based regularizer~\cite{See:2017}, and a beam search with reference mechanism~\cite{Tan:2017}.
We want to investigate the effectiveness of these techniques on sentence summarization, which has not been explored in previous work. 

\vspace{0.05in}
\noindent\textbf{Learning objective.} 
Our training proceeds by minimizing a per-target-word cross-entropy loss function.
A regularization term is applied to the $\boldsymbol\alpha$ matrix. 
Recall that $\alpha_{t,i} \in [0,1]$ measures the interaction strength between the $t$-th output word and the $i$-th input word.
Naturally, we expect a 1-to-1 mapping between the two words. 
The coverage-based regularizer, proposed by See et al.,~\shortcite{See:2017}, encourages this behavior by tracking the historical attention values attributed to the $i$-th input word (up to time step $t$-$1$), denoted by $\widetilde{\alpha}_{t,i} = \sum_{t'=0}^{t-1} \alpha_{t',i}$.
The approach then takes the minimum between $\widetilde{\alpha}_{t,i}$ and $\alpha_{t,i}$, which has the practical effect of forcing $\alpha_{t,i}$ ($\forall t$) to be close to either 0 or 1, otherwise a penalty will be applied.
The regularizer $\Omega$ is defined in Eq.~(\ref{equ:reg_alpha_beta}), where $M$ is the size of the mini-batch, $S$ and $T$ are the lengths of the source and target sequences.
For two-way copy mechanisms, $\boldsymbol\delta$ replaces $\boldsymbol\alpha$ to become the new attention values, we therefore apply regularization to $\boldsymbol\delta$ instead of $\boldsymbol\alpha$.
When the regularizer applies, the objective becomes minimizing ($\mathcal{L} + \Omega$). 
{\medmuskip=1mu
\thinmuskip=1mu
\thickmuskip=1mu
\nulldelimiterspace=2pt
\scriptspace=1pt
\begin{align*}
& \Omega = \lambda \sum_{m=1}^M \frac{1}{T^{(m)}S^{(m)}} \sum_{t=1}^{T^{(m)}} \sum_{i=1}^{S^{(m)}} \Big(\min(\widetilde{\alpha}_{t,i}, \alpha_{t,i})\Big)
\numberthis\label{equ:reg_alpha_beta}
\end{align*}}

\noindent\textbf{Beam search with reference.}
During testing, we employ greedy search to generate system summary sequences.
For the task of summarization, the ground truth summary sequences are usually close to the source texts.
This property can be leveraged in beam search.
Tan et al.,~\shortcite{Tan:2017} describe a beam search with reference mechanism that rewards system summaries that have a high degree of bigram overlap with the source texts.
We describe it in Eq.~(\ref{equ:score_w}), where
where $\mathcal{S}(w)$ denotes the score of word $w$.
$\mathcal{B}(\mathbf{y}_{<t},\mathbf{x})$ measures the number of bigrams shared by the system summary (up to time step $t$-1) and the source text; 
$\{\mathbf{y}_{< t}, w\}$ adds a word $w$ to the end of the system summary. 
The shorter the source text (measured by length $S$), the more weight a shared bigram will add to the score of the current word $w$.
A hyperparameter $\eta$ controls the degree of closeness between the system summary and the source text.
{\medmuskip=1mu
\thinmuskip=1mu
\thickmuskip=1mu
\nulldelimiterspace=2pt
\scriptspace=2pt
\begin{align*}
\mathcal{S}(w) = \log P(w) + \eta \frac{\mathcal{B}(\{\mathbf{y}_{< t}, w\}, \mathbf{x}) - \mathcal{B}(\mathbf{y}_{<t}, \mathbf{x})}{S}
\numberthis\label{equ:score_w}
\end{align*}}

\vspace{-0.3in}
\section{Experiments}
\label{sec:experiments}

We evaluate the proposed structure-infused copy mechanisms for summarization in this section. 
We describe the dataset, experimental settings, baselines, and finally, evaluation results and analysis. 

\subsection{Data Sets}
\label{sec:data}

We evaluate our proposed models on the Gigaword summarization dataset~\cite{Parker:2011,Rush:2015}.
The task is to reduce the first sentence of an article to a title-like summary. 
We obtain dependency parse trees for source sentences using the Stanford neural network parser~\cite{Chen:2014:EMNLP}.
We also use the standard train/valid/test data splits.
Following~\cite{Rush:2015}, the train and valid splits are pruned\footnote{\url{https://github.com/facebookarchive/NAMAS/blob/master/dataset/filter.py}} to improve the data quality. 
Spurious pairs that are repetitive, overly long/short, and pairs whose source and summary sequences have little word overlap are removed. 
No pruning is performed for instances in the test set.
The processed corpus contains 4,018K training instances.
We construct two (non-overlapped) validation sets: ``valid-4096'' contains 4,096 randomly sampled instances from the valid split; it is used for hyperparameter tuning and early stopping. ``valid-2000'' is used for evaluation; it allows the models to be trained and evaluated on pruned instances. 
Finally, we report results on the standard Gigaword test set~\cite{Rush:2015} containing 1,951 instances (``test-1951'').

\subsection{Experimental Setup}

We use the Xavier scheme~\cite{Glorot:2010} for parameter initialization, where weights are initialized using a Gaussian distribution $\mathbf{W}_{i,j} \sim \mathcal{N}(0,\,\sigma)$,  $\smash{\sigma = \sqrt{\frac{2}{n_{in} + n_{out}}}}$; $n_{in}$ and  $n_{out}$ are numbers of the input and output units of the network; biases are set to be 0.
We further implement two techniques to accelerate mini-batch training.
First, all training instances are sorted by the source sequence length and partitioned into mini-batches. 
The shorter sequences are padded to have the same length as the longest sequence in the batch. All batches are shuffled at the beginning of each epoch.
Second, we introduce a variable-length batch vocabulary containing only source words of the current mini-batch and words of the output vocabulary. 
$P(w)$ in Eq.~(\ref{equ:p_w}) only needs to be calculated for words in the batch vocabulary. 
It is magnitudes smaller than a direct combination of the input and output vocabularies.
Finally, our input vocabulary contains the most frequent 70K words in the source texts and summaries.
The output vocabulary contains 5K words by default.
More network parameters are presented in Table~\ref{tab:settings}.

\begin{table}[t]

\begin{minipage}[b]{0.48\textwidth}
\centering
\setlength{\tabcolsep}{6pt}
\renewcommand{\arraystretch}{1.2}
\centering
\begin{small}

\begin{tabular}[t]{|l|l|}
\hline
Input vocabulary size & 70K\\
Output vocabulary size & 5K (default)\\
Dim. of word embeddings & 100\\
Dim. of structural embeddings & 16\\
Num. of encoder/decoder hidden units & 256\\
\hline
Adam optimizer{\scriptsize~\cite{Kingma:2015}} & $lr$ = 1e-4\\
Coeff. for coverage-based regularizer & $\lambda$ = 1\\
Coeff. for beam search with reference & $\eta \approx$ 13.5\\ 
Beam size & $K$ = 5\\
Minibatch size & $M$ = 64\\
Early stopping criterion (max 20 epochs) & valid. loss\\
Gradient clipping{\scriptsize~\cite{Pascanu:2013}} & g $\in$ [-5, 5]\\
\hline
\end{tabular}
\end{small}
\caption{Parameter settings of our summarization system.}
\vspace{0.1in}
\label{tab:settings}

\setlength{\tabcolsep}{8pt}
\renewcommand{\arraystretch}{1.1}
\centering
\begin{small}
\begin{tabular}[t]{|l|ccc|}
\hline
& \multicolumn{3}{c|}{\textbf{Gigaword Valid-2000}}\\
\textbf{System} & \textbf{R-1} & \textbf{R-2} & \textbf{R-L}\\
\hline
\hline
Baseline & 42.48 & {21.34} & 40.18\\
Struct+Input & 42.44 & {21.75} &  {40.46}\\
Struct+Hidden &  {42.88} & {21.81} &  {40.63}\\
Struct+2Way+Word &  \textbf{43.21} & {21.84} &  \textbf{40.86}\\
Struct+2Way+Relation &  {42.83} & \textbf{21.85} &  {40.60}\\
\hline
\end{tabular}
\end{small}
\caption{Results on the Gigaword valid-2000 set (full-length F1).
Models implementing the structure-infused copy mechanisms (``Struct+*'') outperform the baseline.
}
\label{tab:results_valid}

\end{minipage}
\hfill
\begin{minipage}[b]{0.48\textwidth}
\centering

\setlength{\tabcolsep}{2pt}
\renewcommand{\arraystretch}{1.05}
\centering
\begin{scriptsize}
\begin{fontppl}
\begin{tabular}{lp{2.8in}}
\hline
S: & the government filed another round of criminal charges in a widening stock options scandal\\
T: & options scandal widens\\
\hline
\hline
B: & government files more charges in stock options scandal\\
I: & another round of criminal charges in stock options scandal\\
H: & charges filed in stock options scandal\\
W: & another round of criminal charges in stock options scandal\\
R: & government files another round of criminal charges in options scandal\\
\hline
\end{tabular}
\end{fontppl}
\end{scriptsize}
\caption{Example system summaries. `S:' source; `T:' target; `B:' baseline; `I:' Struct+Input; `H:' Struct+Hidden; `W:' 2Way+Word; ``R:'' 2Way+Relation.
``2Way+Relation'' is able to preserve important source relations in the summary, e.g., ``government $\xleftarrow{\vbox{\hbox{\scriptsize nsubj}\vskip-5pt}}$ files,'' ``files $\xrightarrow{\vbox{\hbox{\scriptsize dobj}\vskip-5pt}}$ round,'' and ``round $\xrightarrow{\vbox{\hbox{\scriptsize nmod}\vskip-4pt}}$ charges.''
}
\vspace{0.1in}
\label{tab:output1}

\setlength{\tabcolsep}{2pt}
\renewcommand{\arraystretch}{1.1}
\centering
\begin{scriptsize}
\begin{fontppl}
\begin{tabular}{lp{2.8in}}
\hline
S: & red cross negotiators from rivals north korea and south korea held talks wednesday on emergency food shipments to starving north koreans and agreed to meet again thursday\\
T: & koreas meet in beijing to discuss food aid from south eds\\
\hline
\hline
B: & north korea , south korea agree to meet again\\
I: & north korea , south korea meet again\\
H: & north korea , south korea meet on emergency food shipments\\
W: & north korea , south korea hold talks on food shipments\\
R: & north korea , south korea hold talks on emergency food shipments\\
\hline
\end{tabular}
\end{fontppl}
\end{scriptsize}
\caption{Example system summaries. 
``Struct+Hidden'' and ``2Way+Relation'' successfully preserve salient source words (``emergency food shipments''), which are missed out by other systems. 
We observe that copying ``hold talks'' from the source also makes the resulting summaries more informative than using the word ``meet.''
}
\label{tab:output2}

\end{minipage}
\end{table}

\begin{table}[t]
\begin{minipage}{0.48\textwidth}
\centering

\setlength{\tabcolsep}{4pt}
\renewcommand{\arraystretch}{1.15}
\centering
\begin{small}
\begin{tabular}{|l|ccc|}
\hline
& \multicolumn{3}{c|}{\textbf{Gigaword Test-1951}}\\
\textbf{System} & \textbf{R-1} & \textbf{R-2} & \textbf{R-L} \\
\hline
\hline
ABS{\scriptsize~\cite{Rush:2015}} & 29.55 & {11.32} & 26.42 \\
ABS+{\scriptsize~\cite{Rush:2015}} & 29.76 & {11.88} & 26.96 \\
Luong-NMT{\scriptsize~\cite{Chopra:2016}} & 33.10 & 14.45 & 30.71 \\
RAS-LSTM{\scriptsize~\cite{Chopra:2016}} & 32.55 & {14.70} & 30.03 \\
RAS-Elman{\scriptsize~\cite{Chopra:2016}} & 33.78 & {15.97} & 31.15 \\
ASC+FSC1{\scriptsize~\cite{Miao:2016}} & 34.17 & {15.94} & 31.92 \\
lvt2k-1sent{\scriptsize~\cite{Nallapati:2016}} & 32.67 & {15.59} & 30.64 \\
lvt5k-1sent{\scriptsize~\cite{Nallapati:2016}} & 35.30 & {16.64} & 32.62 \\
Multi-Task{\scriptsize~\cite{Pasunuru:2017}} & 32.75 & {15.35} & 30.82 \\
DRGD{\scriptsize~\cite{Li:2017:DRGD}} & 36.27 & {17.57} & 33.62 \\
\hline
\hline
Baseline (this paper) & 35.43 & 17.49 & 33.39\\
Struct+Input (this paper) & 35.32 & {17.50} & 33.25\\
Struct+2Way+Relation (this paper) & 35.46 & {17.51} & 33.28\\
Struct+Hidden (this paper) & \textbf{35.49} & {17.61} & 33.33\\
Struct+2Way+Word (this paper) & 35.47 & \textbf{17.66} & \textbf{33.52}\\
\hline
\end{tabular}
\end{small}
\caption{Results on the Gigaword test-1951 set (full-length F1). 
Models with structure-infused copy mechanisms (``Struct+*'') perform well. Their R-2 F-scores are on-par with or outperform state-of-the-art published systems. 
}
\label{tab:results_test}

\end{minipage}
\hfill
\begin{minipage}{0.48\textwidth}
\centering

\begin{footnotesize}
\begin{tabular}{|p{2.9in}|}
\hline
\begin{itemize}[topsep=5pt,itemsep=-1pt,itemindent=-0.12in,labelindent=-0.11in,leftmargin=*,before=\vspace{-0.1in},after=\vspace{-0.15in}]
\item[] \textbf{{ABS}} and \textbf{{ABS+}}{\scriptsize~\cite{Rush:2015}} are the first work introducing an encoder-decoder architecture for summarization.

\item[] \textbf{{Luong-NMT}}{\scriptsize~\cite{Chopra:2016}} is a re-implementation of the attentive stacked LSTM encoder-decoder of Luong et al.~\shortcite{Luong:2015}.

\item[] \textbf{{RAS-LSTM}} and \textbf{{RAS-Elman}}{\scriptsize~\cite{Chopra:2016}} describe a convolutional attentive encoder that ensures the decoder focuses on appropriate words at each step of generation.

\item[] \textbf{{ASC+FSC1}}{\scriptsize~\cite{Miao:2016}} presents a generative auto-encoding sentence compression model jointly trained on labelled/unlabelled data. 

\item[] \textbf{{lvt2k-1sent}} and \textbf{{lvt5k-1sent}}{\scriptsize~\cite{Nallapati:2016}} address issues in the attentive encoder-decoder framework, including modeling keywords, capturing sentence-to-word structure, and handling rare words.

\item[] \textbf{{Multi-Task w/ Entailment}}{\scriptsize~\cite{Pasunuru:2017}} combines entailment with summarization in a multi-task setting. 

\item[] \textbf{{DRGD}}{\scriptsize~\cite{Li:2017:DRGD}} describes a deep recurrent generative decoder learning latent structure of summary sequences via variational inference. 
\vspace{5pt}
\end{itemize}\\
\hline
\end{tabular}
\end{footnotesize}
\caption{Existing summarization methods.}
\label{tab:baselines}

\end{minipage}
\end{table}

\subsection{Results}
\label{sec:results}

\textbf{ROUGE results on valid set.}
We first report results on the Gigaword valid-2000 dataset in Table~\ref{tab:results_valid}.
We present R-1, R-2, and R-L scores~\cite{Lin:2004} that respectively measures the overlapped unigrams, bigrams, and longest common subsequences between the system and reference summaries\footnote{w/ ROUGE options: {\fontfamily{fontppl}\selectfont -n 2 -m -w 1.2 -c 95 -r 1000}}.
Our baseline system (``Baseline'') implements the seq2seq architecture with the basic copy mechanism (Eq.~(\ref{equ:h_i_e}-\ref{equ:p_w})).
It is a strong baseline that resembles the pointer-generator networks described in~\cite{See:2017}.
The structural models (``Struct+*'') differ from the baseline only on the structure-infused copy mechanisms.
All models are evaluated without the coverage regularizer or beam search (\S\ref{sec:cov_beam}) to ensure fair comparison.
Overall, we observe that models equipped with the structure-infused copy mechanisms are superior to the baseline, suggesting that combining source syntactic structure with the copy mechanism is effective.
We found that the ``Struct+Hidden'' architecture, which directly concatenates structural embeddings with the encoder hidden states, outperforms ``Struct+Input'' despite that the latter requires more parameters.
``Struct+2Way+Word'' also demonstrates strong performance, achieving 43.21\%, 21.84\%, and 40.86\% F$_1$ scores, for R-1, R-2, and R-L respectively.

\vspace{0.08in}
\noindent\textbf{ROUGE results on test set.}
We compare our proposed approach with a range of state-of-the-art neural summarization systems.
Results on the standard Gigaword test set (``test-1951'') are presented in Table~\ref{tab:results_test}.
Details about these systems are provided in Table~\ref{tab:baselines}.
Overall, our proposed approach with structure-infused pointer networks perform strongly, yielding ROUGE scores that are on-par with or surpassing state-of-the-art published systems.
Notice that the scores on the valid-2000 dataset are generally higher than those of test-1951.
This is because the (source, summary) pairs in the Gigaword test set are not pruned (see \S\ref{sec:data}). 
In some cases, none (or very few) of the summary words appear in the source. 
This may cause difficulties to the systems equipped with the copy mechanism.
The ``Struct+2Way+Word'' architecture that respectively models the semantic and syntactic importance of source words achieves the highest scores.
It outperforms its counterpart of ``Struct+2Way+Relation,'' which seeks to preserve source dependency relations in summaries.
We conjecture that the imperfect dependency parse trees generated by the parser may affect the ``Struct+2Way+Relation'' results. 
However, because the Gigaword dataset does not provide gold-standard annotations for parse trees, we could not easily verify this and will leave it for future work.
In Table~\ref{tab:output1} and~\ref{tab:output2}, we present system summaries produced by various models.

\begin{wraptable}{r}{0.45\textwidth}
\setlength{\tabcolsep}{4pt}
\renewcommand{\arraystretch}{1.1}
\centering
\begin{small}
\begin{tabular}{|l|ccc|}
\hline
\textbf{System} & \textbf{Info.} & \textbf{Fluency} & \textbf{Faithful.} \\
\hline
Struct+Input & 2.9 & 3.3 & 3.0 \\
Struct+2Way+Relation & 3.0 & 3.4 & 3.1\\
Ground-truth Summ. & 3.2 & 3.5 & 3.1\\
\hline
\end{tabular}
\end{small}
\caption{Informativeness, fluency, and faithfulness scores of summaries. They are rated by Amazon turkers on a Likert scale of 1 (worst) to 5 (best). We choose to evaluate Struct+2Way+Relation (as oppose to 2Way+Word) because it focuses on preserving source relations in the summaries. 
}
\label{tab:results_human}
\vspace{-0.1in}
\end{wraptable}

\vspace{0.08in}
\noindent\textbf{Linguistic quality.}
To further gauge the summary quality, we hire human workers from the Amazon Mechanical Turk platform to rate summaries on a Likert scale of 1 to 5 according to three criteria~\cite{Zhang:2017}:
\emph{fluency} (is the summary grammatical and well-formed?), \emph{informativeness} (to what extent is the meaning of the original sentence preserved in the summary?), and \emph{faithfulness} (is the summary accurate and faithful to the original?).
We sample 100 instances from the test set and employ 5 turkers to rate each summary; their averaged scores are presented in Table~\ref{tab:results_human}.
We found that ``Struct+2Way+Relation'' outperforms ``Struct+Input'' on all three criteria.
It also compares favorably to ground-truth summaries on ``fluency'' and ``faithfulness.''
On the other hand, the ground-truth summaries, corresponding to article titles, are judged as less satisfying according to human raters.

\vspace{0.08in}
\noindent\textbf{Dependency relations.}
We investigate the source dependency relations preserved in the summaries in Table~\ref{tab:stats}. 
A source relation is considered preserved if both its words appear in the summary.
We observe that the models implementing structure-infused copy mechanisms (e.g., ``Struct+2Way+Word'') are more likely to preserve important dependency relations in the summaries, including {\fontppl\small nsubj}, {\fontppl\small dobj}, {\fontppl\small amod}, {\fontppl\small nmod}, and {\fontppl\small nmod:poss}.
Dependency relations that are less important ({\fontppl\small mark}, {\fontppl\small case}, {\fontppl\small conj}, {\fontppl\small cc}, {\fontppl\small det}) are less likely to be preserved.
These results show that our structure-infused copy mechanisms can learn to recognize the importance of dependency relations and selectively preserve them in the summaries.

\begin{table*}
\setlength{\tabcolsep}{6pt}
\renewcommand{\arraystretch}{1.1}
\centering
\begin{small}
\begin{tabular}{|l|ccccc|ccccc|}
\hline
\textbf{System} & \textbf{nsubj} & \textbf{dobj} & \textbf{amod} & \textbf{nmod} & \textbf{nmod:poss} & \textbf{mark} & \textbf{case} & \textbf{conj} & \textbf{cc} & \textbf{det}\\
\hline
\hline
Baseline & 7.23 & 12.07 & 20.45 & 8.73 & 12.46 & 15.83 & 14.84 & 9.72 & 5.03 & 2.22\\
\hline
Struct+Input & 7.03 & 11.72 & 19.72 & \textbf{9.17}$\uparrow$ & {12.46} & 15.35 & 14.69 & 9.55 & 4.67 & 1.97\\
Struct+Hidden & \textbf{7.78}$\uparrow$ & \textbf{12.34}$\uparrow$ & \textbf{21.11}$\uparrow$ & \textbf{9.18}$\uparrow$ & \textbf{14.86}$\uparrow$ & 14.93 & \textbf{15.84}$\uparrow$ & 9.47 & 3.93 & \textbf{2.65}$\uparrow$\\
Struct+2Way+Word & \textbf{7.46}$\uparrow$ & \textbf{12.69}$\uparrow$ & \textbf{20.59}$\uparrow$ & \textbf{9.03}$\uparrow$ & \textbf{13.00}$\uparrow$ & 15.83 & 14.43 & 8.86 & 3.48 & 1.91\\
Struct+2Way+Relation & \textbf{7.35}$\uparrow$ & \textbf{12.07}$\uparrow$ & \textbf{20.59}$\uparrow$ & 8.68 & \textbf{13.47}$\uparrow$ & 15.41 & 14.39 & 9.12 & 4.30 & 1.89\\
\hline
\end{tabular}
\end{small}
\caption{Percentages of source dependency relations (of various types) preserved in the system summaries. 
}
\label{tab:stats}
\end{table*}

\begin{table}[t]
\begin{minipage}[b]{0.48\textwidth}
\centering

\includegraphics[width=2.7in]{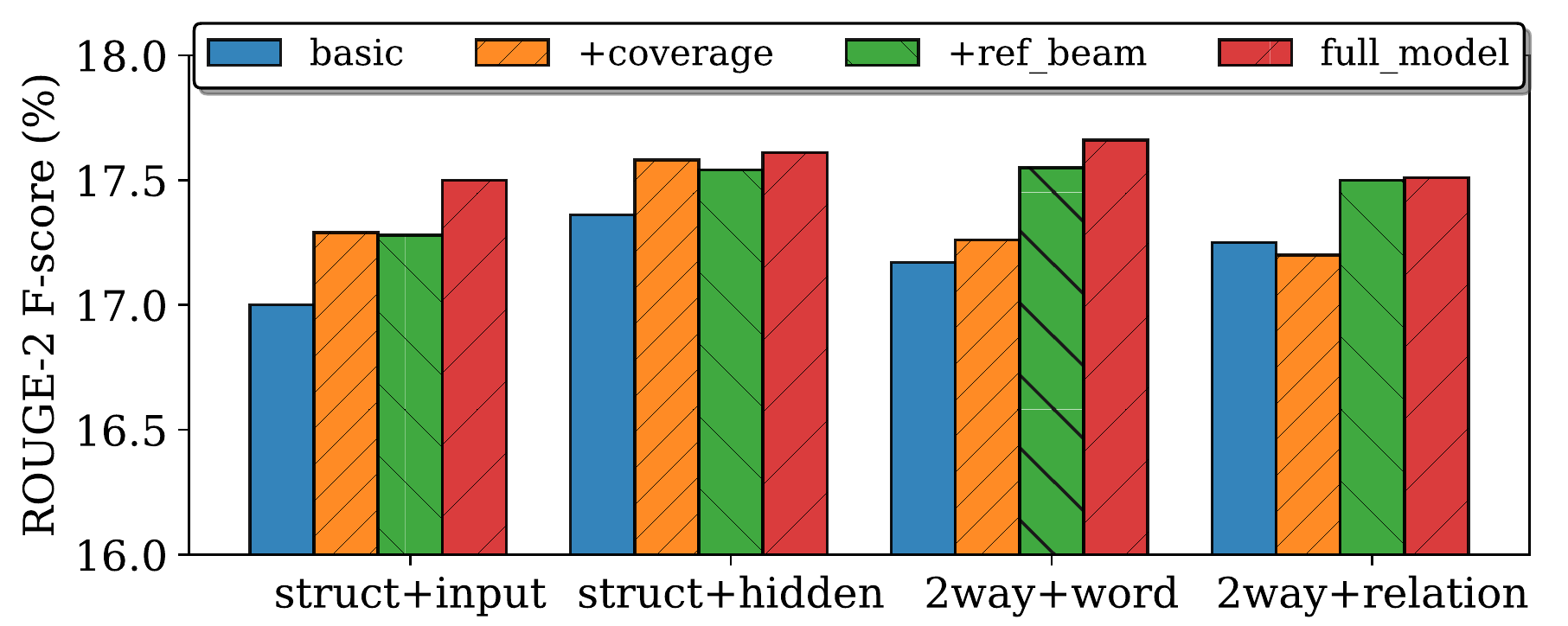}
\caption{Effects of applying the coverage regularizer and the reference beam search to structural models, evaluated on test-1951. Combining both yields the highest scores.
}
\label{fig:cov_beam}

\end{minipage}
\hfill
\begin{minipage}[b]{0.48\textwidth}
\centering

\setlength{\tabcolsep}{5pt}
\renewcommand{\arraystretch}{1.1}
\centering
\begin{small}
\begin{tabular}{|l|c|c|c|c|}
\hline
\textbf{$|V|$} & \textbf{R-2} & \textbf{Train Speed} & \textbf{InVcb} & \textbf{InVcb+Src}\\
\hline
1K & 13.99 & 2.5h/epoch & 60.57 & 76.04\\
2K & 15.35 & 2.7h/epoch & 69.71 & 80.72\\
5K & 17.25 & 3.2h/epoch & 79.98 & 86.51\\
10K & 17.62 & 3.8h/epoch & 88.26 & 92.18\\
\hline
\end{tabular}
\end{small}
\caption{Results of the ``Struct+2Way+Relation'' system trained using output vocabularies of various sizes ($|V|$), evaluated on test-1951 w/o coverage or ref\_beam.
The training speed is calculated as the elapsed time (hours) per epoch, tested on a GTX 1080Ti GPU card. 
}
\label{tab:vocab_size}

\end{minipage}
\vspace{-0.1in}
\end{table}

\vspace{0.08in}
\noindent\textbf{Coverage and reference beam.}
In Figure~\ref{fig:cov_beam}, we investigate the effect of applying the coverage regularizer (``coverage'') and reference-based beam search (``ref\_beam'') (\S\ref{sec:cov_beam}) to our models. 
The coverage regularizer is applied in a second training stage, where the system is trained for an extra 5 epochs with coverage and the model yielding the lowest validation loss is selected.
Both coverage and ref\_beam can improve the system performance. Our observation suggests that ref\_beam is an effective addition to shorten the gap between different systems. 

\vspace{0.08in}
\noindent\textbf{Output vocabulary size.}
Finally, we investigate the impact of the output vocabulary size on the summarization performance in Table~\ref{tab:vocab_size}.
All our models by default use an output vocabulary of 5K words in order to make the results comparable to state-of-the-art-systems. 
However, we observe that there is a potential to further boost the system performance (17.25$\rightarrow$17.62 R-2 F$1$-score, w/o coverage or ref\_beam) if we had chosen to use a larger vocabulary (10K) and can endure a slightly longer training time (1.2x).
In Table~\ref{tab:vocab_size}, we further report the percentages of reference summary words covered by the output vocabulary (``InVcb'') and covered by either the output vocabulary or the source text (``InVcb+Src'').
The gap between the two conditions shortens as the size of the output vocabulary is increased.

\section{Conclusion}
\label{sec:conclusion}

In this paper, we investigated structure-infused copy mechanisms that combine source syntactic structure with the copy mechanism of an abstractive summarization system.
We compared various system architectures and showed that our models can effectively preserve salient source relations in summaries. 
Results on benchmark datasets showed that the structural models are on-par with or surpass state-of-the-art published systems.



\bibliographystyle{acl}
\bibliography{summ,abs_summ,fei}

\end{document}